\documentclass[10pt,twocolumn,letterpaper]{article}

\usepackage{cvpr}              
\usepackage{multirow}

\definecolor{cvprblue}{rgb}{0.21,0.49,0.74}
\usepackage[pagebackref,breaklinks,colorlinks,allcolors=cvprblue]{hyperref}

\def\paperID{13} 
\def\confName{CVPR}
\def\confYear{2025}

\title{HDC: Hierarchical Distillation for Multi-level Noisy Consistency in Semi-Supervised Fetal Ultrasound Segmentation}

\author{Tran Quoc Khanh Le $^{1,4*}$ \quad Nguyen Lan Vi Vu$^{2, 4}$\thanks{Equal Contribution.} \quad  Ha-Hieu Pham$^{3,4}$ \quad  Xuan-Loc Huynh$^{5}$ \\ 
Tien-Huy Nguyen$^{1,4}$ \quad Minh Huu Nhat Le$^{6}$ \quad Quan Nguyen$^{7}$ \quad Hien D. Nguyen$^{1,4,8}$\thanks{Corresponding author.} \vspace{1.5mm}\\ 
$^{1}$University of Information Technology, Ho Chi Minh city, Vietnam \\
$^{2}$Ho Chi Minh University of Technology, Ho Chi Minh city, Vietnam \\
$^{3}$University of Science, Ho Chi Minh city, Vietnam \\
$^{4}$Vietnam National University, Ho Chi Minh city, Vietnam \\
$^{5}$Boston University, MA, USA \\
$^{6}$Taipei Medical University, Taipei, Taiwan \\
$^{7}$Posts and Telecommunications Institute of Technology, Hanoi, Vietnam \\
$^{8}$New Mexico State University, Las Cruces, NM, USA}
\begin{document}
\maketitle
\begin{abstract}
Transvaginal ultrasound is a critical imaging modality for evaluating cervical anatomy and detecting physiological changes. However, accurate segmentation of cervical structures remains challenging due to low contrast, shadow artifacts, and indistinct boundaries. While convolutional neural networks (CNNs) have demonstrated efficacy in medical image segmentation, their reliance on large-scale annotated datasets presents a significant limitation in clinical ultrasound imaging. Semi-supervised learning (SSL) offers a potential solution by utilizing unlabeled data, yet existing teacher-student frameworks often encounter confirmation bias and high computational costs. In this paper, a novel semi-supervised segmentation framework, called HDC, is proposed incorporating adaptive consistency learning with a single-teacher architecture. The framework introduces a hierarchical distillation mechanism with two objectives: Correlation Guidance Loss for aligning feature representations and Mutual Information Loss for stabilizing noisy student learning. The proposed approach reduces model complexity while enhancing generalization. Experiments on fetal ultrasound datasets, FUGC and PSFH, demonstrate competitive performance with reduced computational overhead compared to multi-teacher models.
\end{abstract}

\section{Introduction}
\label{sec:intro}

\begin{figure}[t]
    \centering
    \includegraphics[width=1\linewidth]{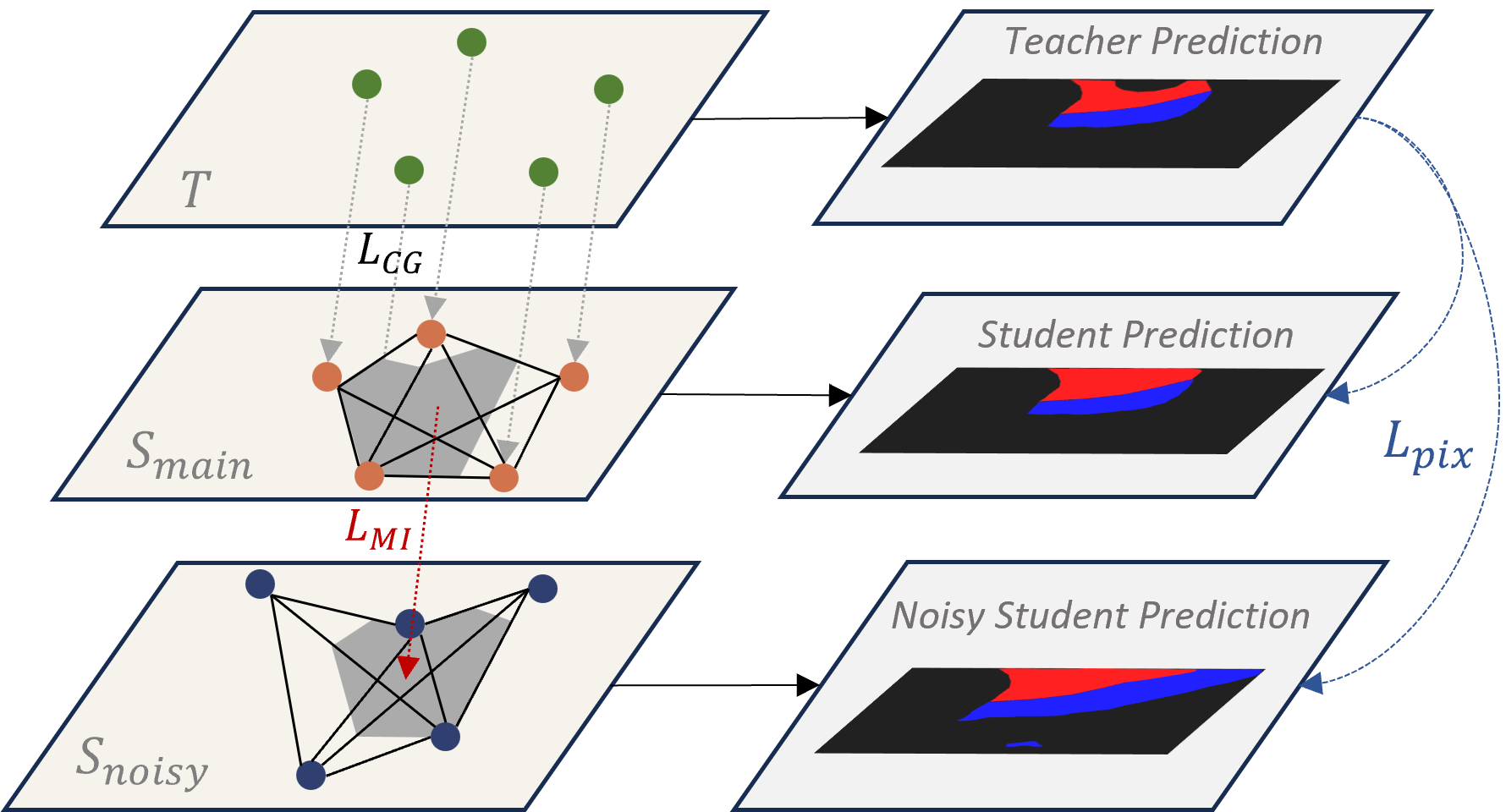}

    \caption{An intuitive overview of our Hierarchical Distillation and Consistency scheme.}
    \label{fig:overview}
\end{figure}

Transvaginal ultrasound is a widely used imaging technique for assessing cervical anatomy, providing essential insights into muscle structure and function. Accurate segmentation of cervical ultrasound images is crucial for evaluating physiological changes and guiding clinical decision-making. However, the segmentation process is challenging due to common ultrasound imaging issues such as shadow artifacts, low contrast, and indistinct boundaries \cite{fugc2025,tran2022segmentation}.

Convolutional Neural Networks (CNNs) have demonstrated strong performance in medical image segmentation by capturing local spatial features. However, traditional supervised learning approaches that rely on CNNs require large-scale annotated datasets to be generalized effectively \cite{nguyen2024fa}. In medical imaging, particularly in transvaginal ultrasound, obtaining pixel-wise labeled data is both time-consuming and costly, limiting the scalability of supervised CNN models \cite{pham2023robust}. Moreover, CNNs primarily focus on local dependencies, which may lead to segmentation errors such as over-segmentation or under-segmentation, particularly in complex anatomical structures.

To overcome these limitations, semi-supervised learning (SSL) has emerged as a promising approach by leveraging both labeled and unlabeled data to improve segmentation performance. Among various SSL frameworks, the Mean Teacher \cite{tarvainen2017mean} model has gained significant attention that ensures smooth training dynamics and improves student and teacher prediction consistency. Recent advancements, such as Dual Teacher \cite{guo2023dual}, introduce multiple teacher networks that periodically switch during training to enhance model robustness and mitigate confirmation bias. These techniques highlight the promise of multi-teacher SSL frameworks but add complexity or intricate update mechanisms.

\begin{figure*}[t]
	\centering
	\includegraphics[width=1\linewidth]{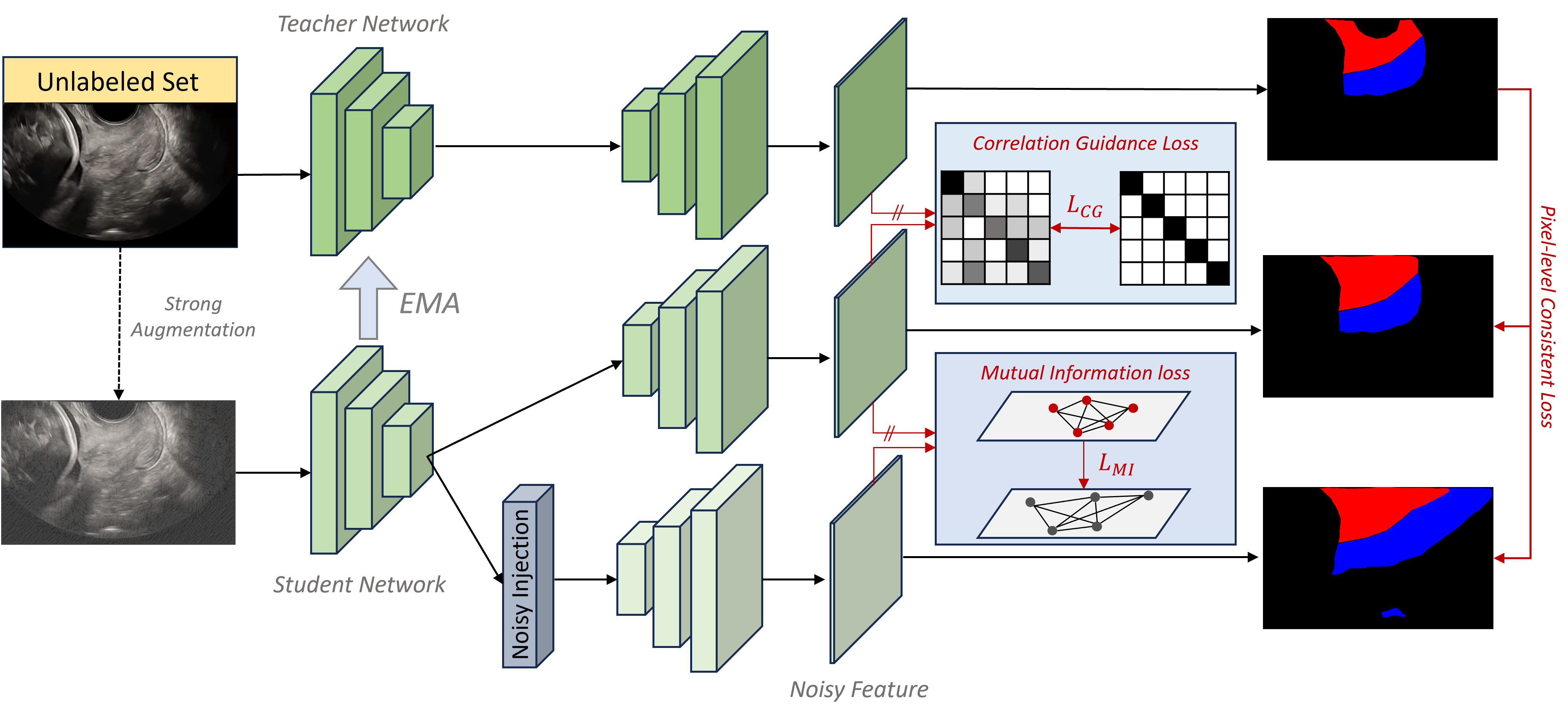}
    \caption{Our proposed semi-supervised pipeline for transvaginal ultrasound segmentation. The teacher generates pseudo-labels to guide the student, which processes both raw and noisy inputs. The dashed arrow (- - -) represents strong augmentation, the double slash arrow (//) denotes feature comparisons for $L_{CG}$ and $L_{MI}$, black arrows indicate forward propagation, and red arrows represent pixel-level consistency loss.}
	\label{fig: 1}
\end{figure*}

Existing dual-teacher and multi-student methods \cite{guo2023dual, zhao2024alternate}, despite their advantages, often suffer from high computational costs due to the need for training and maintaining multiple models. Some approaches address confirmation bias by incorporating strong-weak augmentation strategies or contrastive learning mechanisms. While these techniques can enhance segmentation performance, they frequently demand careful hyperparameter tuning and extensive computational resources, making them less practical for real-world medical imaging applications.


To address the challenges of semi-supervised medical image segmentation—particularly confirmation bias and computational inefficiency—this work introduces a novel framework that combines adaptive consistency training with a single-teacher architecture. Unlike conventional approaches that rely on multiple teacher-student models \cite{guo2023dual, zhao2024alternate}, our method enhances training efficiency and generalization through structured noise injection and carefully designed objective functions .Most existing teacher-student methods focus primarily on transferring knowledge through pseudo-labeling based solely on output predictions. 

In contrast, we propose a hierarchical distillation mechanism that enriches the learning signal and improves feature alignment, inspired by theoretical insights from prior works on knowledge distillation \cite{zbontar2021barlow, miles2021information}. This is achieved through two novel loss functions: \textbf{(1) Correlation Guidance Loss}, which encourages consistency between the feature representations of the teacher and the main student branch, and \textbf{(2) Mutual Information Loss}, which aligns the representations between the main student and a noisy student branch to enhance intra-batch stability. These mechanisms are designed to bridge the feature distribution gap between teacher and student networks while stabilizing student representations in the presence of noise. HDC is evaluated on two challenging fetal ultrasound datasets, FUGC and PSFH \cite{fugc2025, lu2022jnu}, where it achieves \textbf{state-of-the-art} segmentation performance with significantly reduced computational overhead compared to existing methods under various evaluation metrics.

\section{Related Works}
\label{sec:relatedworks}
\textbf{Semi-Supervised Learning (SSL). }
The core challenge of semi-supervised learning is to maximize model accuracy while minimizing human annotation effort by effectively leveraging abundant unlabeled data in conjunction with a small set of high-quality labeled samples. Foundational works in SSL can be categorized into two dominant branches: Pseudo-labeling \cite{rizve2021defense,cascante2021curriculum} and consistency regularization \cite{Ouali_2020_CVPR,ngo2024dual, nguyen2024blurry}. Pseudo-labeling follows an iterative training process. The most up-to-date optimized model generates pseudo-labels from unlabeled data and combines them with labeled samples to refine prediction over time. On the other hand, consistency regularization-based methods rely on the smoothness assumption \cite{learning2006semi}, stemming from the goal of training models that remain invariant to various perturbations applied to either the input data or the models themselves. FixMatch \cite{sohn2020fixmatch} integrates the strength of both strategies by employing a weak-to-strong consistency mechanism. In this research, pseudo labels are derived from weakly augmented samples and used to supervise the training of their strongly augmented counterparts. FlexMatch \cite{zhang2021flexmatch} takes into account different learning difficulties across categories and creates different thresholds for each class. 

\noindent \textbf{Semi-Supervised Semantic Segmentation (S4). } Benefits from the advances in deep neural networks \cite{lee2013pseudo,vesely2013semi,hong2015decoupled} and robust SSL algorithms, many works adopt them to build powerful semantic segmentation baselines \cite{liang2023logic,na2023switching}. In medical image analysis, SSL also advances medicine significantly, with many noticeable works \cite{nguyen2024blurry, luu2025semi, nguyen5109180mv}. On top of these fundamental designs, Mean-Teacher-based methods have achieved remarkable results across benchmarks. The Mean Teacher (MT) framework \cite{tarvainen2017mean} introduces a teacher-student architecture in which the teacher model, serving as a stable version of the student, is updated using the exponential moving average (EMA) of the student's weights to guide the student during training. Building upon MT, PS-MT \cite{liu2022perturbed} proposes a stricter teacher that prevents model collapse. AD-MT \cite{zhao2024alternate} employs a dual-teacher framework with different perspectives in an adaptive switching manner to mitigate confirmation bias. On the other hand, co-training has also shown promising results in S4 by leveraging multi-view references to improve model perception and enhance pseudo-label reliability. Cross-pseudo supervision (CPS) \cite{chen2021-CPS} enforces consistency between the outputs of two networks initialized with different weights to foster diverse feature learning. CCVC \cite{wang2023conflict} suggests learning richer semantic information from competing predictions by pushing the feature extractor outputs of two networks apart, thus mitigating the detrimental effects of identical coupling problems in CPS.

\noindent \textbf{Knowledge Distillation (KD). } Originally introduced by Hinton et al. \cite{hinton2015distilling}, this technique enables the transfer of knowledge from a large, complex teacher model to a smaller student model, enabling efficient deployment while maintaining performance. KD plays important roles in opening new approaches for many fields such as multi-modal tasks\cite{le2023enhancing, nguyen2024improving, tran2025mlg2net}. Originally proposed as a method where the student mimics the teacher’s soft predictions, KD has since evolved into multiple paradigms, each capturing different aspects of knowledge transfer. KD methods are commonly categorized into logits-based \cite{hinton2015distilling,jin2023multi}, feature-based \cite{chen2022knowledge,chen2021distilling,guo2023class}, and relation-based approaches \cite{huang2022knowledge,mei2023conditional}. Response-based distillation aligns the student’s predictions with the teacher’s soft outputs, often using Kullback-Leibler \cite{van2014renyi} divergence. Beyond classical approaches, recent works have explored alternative formulations grounded in self-supervision and information theory \cite{miles2021information}, aiming for a more structured and effective knowledge transfer. Some methods adopt orthogonal projection techniques \cite{miles2024vkd} to distill knowledge more efficiently, while others focus on activation gradient alignment to better capture the loss landscape \cite{srinivas2018knowledge}. 

\section{Method}

In semi-supervised segmentation, the training set consists of a smaller labeled subset, $\mathbf{D}_l = \{(x_m^l, y_m^l)\}_{m=1}^{M}$, and a significantly larger unlabeled subset, $\mathbf{D}_u = \{(x_n^u)\}_{n=1}^{N}$, where $N \gg M$. Each fetal ultrasound image, $x_m^l$ or $x_n^u$, is represented as a tensor of size $3 \times H \times W$. Labeled images $x_m^l$ are paired with their corresponding ground-truth segmentation masks $y_m^l$.  

\subsection{Theoretical Insights}




\noindent \textbf{Rényi’s Entropy \cite{renyi1961measures}} is a generalized measure of uncertainty that extends Shannon entropy by introducing a tunable parameter $\alpha$, which adjusts the sensitivity to different probability distributions. The Rényi entropy of order $\alpha$ for a discrete random variable $X$ with probability mass function $p(x)$ is defined as:
\begin{equation}
    H_\alpha (X) = \frac{1}{1 - \alpha} \log \left( \sum_{i=1}^{n} p_i^\alpha \right)
\end{equation}
When $\alpha = 1$ we get the more familiar Shannon entropy $H_1(X) = \lim_{\alpha \to 1} H_\alpha(X) = -  \sum_{i=1}^{n} p_i \log_2 p_i $.

\noindent \textbf{Mutual Information (MI) \cite{cover1999elements}} measures the amount of information shared between two random variables. It represents the reduction in uncertainty of one variable given knowledge of the other. Given two random variables $X$ and $Y$ with a joint probability mass function $ p(x, y) $ and marginal distributions $ p(x) $ and $ p(y) $. The mutual information $ I(X; Y) $ is defined as the Kullback-Leibler divergence between the joint distribution and the product of the marginals: 
\begin{equation}
\begin{split}
    I(X; Y) &= \sum_{x \in \mathcal{X}} \sum_{y \in \mathcal{Y}} p(x, y) \log \frac{p(x, y)}{p(x)p(y)} \\
    &= - \sum_{x,y} p(x, y) \log p(x) + \sum_{x,y} p(x, y) \log p(x|y) \\
    &= H(X) - H(X|Y)
\end{split}
\end{equation}
Applying the relationship between joint entropy and relative entropy, $ H(X, Y) = H(Y) + H(X|Y) $, we have
\label{eq:3}
\begin{equation}
    I(X; Y) = H(X) + H(Y) - H(X, Y).
\end{equation}
\noindent \textbf{Feature-based Knowledge Distillation \cite{gou2021knowledge}}  uses latent feature representations of a high-capacity teacher model as the knowledge to supervise the training of student model. Rather than distilling knowledge from intermediate layers, we extract the final feature representations just before the fully connected layer. This preserves the structural information of the input while reducing the adverse effects of inductive biases in cross-architecture distillation.

\noindent We model the structural information through a Gram matrix $K \in \mathbb{R}^{b \times b}$, where $b$ represents the batch size. This Gram matrix encodes the pairwise similarity among all features in a batch:
\begin{equation}
K_{ij} = k(\mathbf{Z}_i, \mathbf{Z}_j) = \langle \mathbf{Z}_i, \mathbf{Z}_j \rangle_{\mathcal{H}},
\end{equation}
where $\mathbf{Z} \in \mathbb{R}^{b \times d}$ corresponds to either teacher or student features with dimension $d$, and $\mathcal{H}$ is a Hilbert space implicitly defined by the positive definite real valued kernel $k$, which can be selected as polynomial kernel or radial basis function kernel.

\noindent Traditional Rényi entropy is defined on probability distributions, which are often difficult to estimate in practice, especially in high-dimensional spaces where explicit density estimation is challenging.  By normalizing $K$ to have a trace of 1 and interpreting its eigenvalues $\lambda(K)$ as an implicit distribution over the feature space, we extend Rényi entropy to structured data. The matrix-based Rényi entropy is then formulated as:
\begin{equation}
H_\alpha(K) = \frac{1}{1 - \alpha} \log_2 \left( \sum_{i=1}^{n} \lambda_i(K^\alpha) \right),
\end{equation}
where $\lambda_i(K)$ are the eigenvalues of $K$, and $\alpha$ is a tunable parameter.

\subsection{Overall Training Scheme}

Our overall workflow is illustrated in Figure \ref{fig: 1}. The proposed HDC framework comprises a teacher network, denoted as $\mathcal{T}$ and a single-encoder, dual-decoder student network. The student's main decoder ($\mathcal{D}_s^{1}$) learns directly from the teacher’s representations and updates the teacher’s weights via EMA. Meanwhile, the noisy decoder ($\mathcal{D}_s^{2}$) is trained with adversarial noise to enhance the robustness and generalization of the shared encoder $\mathcal{E}_s$ through backward propagation.
 
\noindent We propose a multi-level noisy Mean Teacher framework, where noise is introduced at 3 different levels. At the lowest level, weak augmentations, including random flips, rotations, and transpositions, are applied to the teacher’s input. Moving to a stronger augmentation level, the student receives intensive-based transformations in unsupervised phase, such as color jittering, auto-contrast and Gaussian noise with a random distortion strength, which enlarges the recognition space and prevents overfitting to the teacher’s predictions.    

\noindent At the highest noise level, structured perturbations are injected into the the student’s noisy decoder. Specifically, we adopt F-noise from \cite{Ouali_2020_CVPR}, where uniform noise $\mathcal{N} \sim \mathcal{U}(-0.3, 0.3)$ is sampled with the same size of the encoder’s output. The perturbed encoder's feature representation is then computed as $\tilde{\mathbf{Z_s}} = (\mathbf{Z}_s \odot \mathcal{N}) + \mathbf{Z}_s$.

 In supervised phase, the labeled input $ x^l $ is transformed into weak augmented versions, then the labels $y^l$ guide both student's decoders outputs:
 \begin{equation}
     p^1 = \mathcal{D}_s^1(\mathcal{E}_s(\mathcal{A}_w(x^l))), \quad p^2 = \mathcal{D}_s^2 (\text{\textit{Noise}} (\mathcal{E}_s(\mathcal{A}_w(x^l)))) ,
 \end{equation}
 we eliminate the Softmax term applied on $p^k$ for efficient, the supervised loss is then computed as:
\begin{equation}
    \mathcal{L}_{\text{sup}} = \sum_{k=1}^{2}
    \frac{1}{M} \sum_{i=1}^{M} \frac{1}{H \cdot W} \sum_{j=1}^{H \cdot W} 
    \mathcal{L}_{\text{ce}}(p_{ij}^{k}, y_{ij}^{l}),
\end{equation}
where $ M $ is the number of labeled images, $ \mathcal{L}_{\text{ce}} $ is the pixel-wise cross-entropy loss, $ p_{ij}^{k} $ is the student's decoder prediction, and \textit{Noise} represents noise injection applied to the noisy student's decoder. The details of the unsupervised phase will be presented in the following sections.


\subsection{Hierarchical Distillation for Multi-Level Noisy Consistency}

In the unsupervised phase, the overall training objective consists of three primary loss functions: Correlation Guidance Loss, which aligns the teacher and student's main decoder representations; Mutual Information Loss, which maximizes information sharing between the two student decoders; and Pixel-level Consistency Loss, which ensures consistency between the probability outputs of the teacher and student models (Figure \ref{fig:overview}).

\subsubsection{Correlation Guidance Loss}

To ensure that the student’s learned features are structurally aligned with the teacher’s representations, we employ a correlation guidance loss. This loss is inspired by the concept of cross-correlation in self-supervised learning \cite{zbontar2021barlow}, where highly correlated representations indicate strong knowledge transfer.

\noindent Given a batch of normalized final feature representations from the student $\mathbf{Z}_s \in \mathbb{R}^{b \times d}$ and the teacher $\mathbf{Z}_t \in \mathbb{R}^{b \times d}$, where $b$ is the batch size, we compute the cross-correlation matrix:
\begin{equation}
    \mathcal{C}_{st} = \frac{\mathbf{Z}_s^T \mathbf{Z}_t}{b}
\end{equation}
 A perfect correlation between the student and teacher representations is achieved when the diagonal entries $(C_{st})_{ii}$ are all equal to 1. To enforce this, we define the correlation loss as:
\begin{equation}
    \mathcal{L}_{\text{CG}} = \log_2 \sum_{i=1}^{d} ((\mathcal{C}_{st})_{ii} - 1)^{2\alpha},
\end{equation}
where $\alpha$ is a hyperparameter that controls the penalty for deviations from perfect correlation. A larger $\alpha$ enforces stronger alignment between the student and the teacher representations. For computational efficiency, we set $\alpha = 2$.

\subsubsection{Mutual Information Loss}

To further enhance knowledge transfer, we move beyond feature-wise alignment and incorporate intra-batch consistency through a mutual information loss, originally proposed in \cite{miles2021information}. While the correlation loss ensures that the student’s main decoder is well-aligned with the teacher, it does not explicitly enforce consistency between the student’s two decoders. To address this, we establish a hierarchical distillation scheme, where knowledge is not only transferred from the teacher to the main decoder but also propagated from the main decoder to the noisy one. This structured flow of information ensures that both decoders develop robust representations while simultaneously allowing the shared encoder to learn complementary features from both branches.

\noindent Given the Gram matrices of the student’s main decoder, $K^1$ and the noisy decoder, $K^2$, we compute their Hadamard product:
\begin{equation}
    K^{12} = K^1 \circ K^2,
\end{equation}
which $K^{12}$ 
  is normalized to ensure unit trace to preserve scale invariance. The mutual information loss is then formulated as:
\begin{equation}  
    \mathcal{L}_{\text{MI}} = -I(K^1, K^2) = H_\alpha(K^{12}) - H_\alpha(K^2),  
\end{equation}  
where $ H_\alpha $ denotes the entropy term parameterized by $\alpha$. Compared to Equation \ref{eq:3}, the term $ K^1 $ is omitted since we stop the gradient of the main decoder's output at this stage. Maximizing the multual information encourages the student's decoders to preserve the intra-batch relationships for transfering structured knowledge across perturbation levels.

\subsubsection{Pixel-Level Consistency Loss}

To enforce fine-grained alignment between the teacher’s predictions and the student's decoders, we introduce a pixel-level consistency loss. Unlike the correlation and mutual information losses, which focus on feature representations, this loss directly minimizes the discrepancy between the teacher’s predicted outputs and the outputs of both student decoders.

Let $\hat{y} = \mathcal{T}(\mathcal{A}_w(x^u))$ denote the teacher's predicted probability map, while $p^1_u$ and $p^2_u$ represent the outputs of the student's main and noisy decoders, respectively, given strongly augmented inputs $\mathcal{A}_s(x)$. To enforce consistency between the student's predictions and the teacher’s guidance, we define the pixel-level consistency loss as:  
\begin{equation}
    \mathcal{L}_{\text{pix}} = \| p^1_u - \hat{y} \|^2_2 + \| p^2_u - \hat{y} \|^2_2
\end{equation}
This loss ensures that both decoders produce outputs that remain close to the teacher’s predictions, thereby enforcing stable predictions while preserving pixel-wise alignment.

\subsection{Final Objective Function}

The final loss function combines supervised loss together with knowledge distillation losses:
\begin{equation}
    \mathcal{L}_{total} = \mathcal{L}_{\text{sup}} + \beta_{\text{CG}} \mathcal{L}_{\text{CG}} + \beta_{\text{MI}} \mathcal{L}_{\text{MI}} +  \mathcal{L}_{\text{pix}},
\end{equation}
where $\beta_{\text{CG}}$ and $\beta_{\text{MI}}$ are weighting hyper-parameters regulate the interplay between correlation guidance from the teacher and the student's mutual information preservation.

\begin{table*}[h]
\centering
\renewcommand{\arraystretch}{1.15} 
\setlength{\tabcolsep}{3pt} 
\tiny
\resizebox{\textwidth}{!}{%
\begin{tabular}{|c|l|p{17pt}p{17pt}|p{17pt}p{17pt}|p{17pt}p{17pt}|}
\hline
\multirow{2}{*}{Backbone} & \multirow{2}{*}{Method} & \multicolumn{2}{c|}{40/450} & \multicolumn{2}{c|}{20/470} & \multicolumn{2}{c|}{10/480} \\ \cline{3-8}
                          &                          & \textbf{DSC $\uparrow$} & \textbf{HD $\downarrow$}   & \textbf{DSC $\uparrow$} & \textbf{HD $\downarrow$}   & \textbf{DSC $\uparrow$} & \textbf{HD $\downarrow$}   \\ 
\hline

\multirow{9}{*}{ResNet50}  
& SupOnly & 0.8360 & 50.91 & 0.8012 & 59.02 & 0.7732 & 66.03 \\ 
& Mean Teacher \hfill\textcolor{gray}{(NeurIPS'17)} & 0.8602 & 49.12 & 0.8372 & 53.15 & 0.8177 & 60.20 \\ 
& FixMatch  \hfill\textcolor{gray}{(NeurIPS'20)}    & 0.8666 & \textbf{38.82} & 0.8347 & 54.14 & 0.8192 & 59.81 \\ 
& CPS  \hfill\textcolor{gray}{(CVPR'21)} & 0.8590 & 43.12 & 0.8321 & 56.08 & 0.8130 & 61.00 \\ 
& PS-MT \hfill\textcolor{gray}{(CVPR'22)} & 0.8621 & 47.00 & 0.8401 & 51.99 & 0.8213 & 59.37 \\ 
& CCVC  \hfill\textcolor{gray}{(CVPR'23)} & 0.8597 & 41.09 & 0.8325 & 55.78 & 0.8165 & 60.42 \\ 
& Dual Teacher  \hfill\textcolor{gray}{(NeurIPS'23)} & 0.8672 & 42.70 & 0.8413 & 51.38 & 0.8203 & 59.55 \\ 
& AD-MT \hfill\textcolor{gray}{(ECCV'24)} & 0.8700 & 39.99 & 0.8427 & 50.66 & 0.8230 & 59.04 \\ 
& \textbf{HDC} & \textbf{0.8704} & 38.97 & \textbf{0.8503} & \textbf{48.73} & \textbf{0.8299} & \textbf{57.82} \\ 
\hline

\multirow{9}{*}{ResNet101}  
& SupOnly & 0.8542 & 39.17 & 0.8137 & 57.45 & 0.7798 & 63.89 \\ 
& Mean Teacher \hfill\textcolor{gray}{(NeurIPS'17)} & 0.8862 & 47.40 & 0.8298 & 51.58 & 0.8201 & 59.59 \\ 
& FixMatch  \hfill\textcolor{gray}{(NeurIPS'20)}    & 0.8874 & \textbf{33.36} & 0.8402 & 52.11 & 0.8221 & 59.21 \\ 
& CPS  \hfill\textcolor{gray}{(CVPR'21)} & 0.8739 & 38.20 & 0.8398 & 51.02 & 0.8169 & 60.58 \\ 
& PS-MT \hfill\textcolor{gray}{(CVPR'22)} & 0.8870 & 40.34 & 0.8420 & 56.25 & 0.8234 & 58.97 \\ 
& CCVC  \hfill\textcolor{gray}{(CVPR'23)} & 0.8791 & 36.67 & 0.8409 & 48.86 & 0.8199 & 59.62 \\ 
& Dual Teacher  \hfill\textcolor{gray}{(NeurIPS'23)} & 0.8898 & 38.86 & 0.8498 & 48.63 & 0.8242 & 58.83 \\ 
& AD-MT \hfill\textcolor{gray}{(ECCV'24)}  & 0.8910 & 36.92 & 0.8507 & 46.78 & 0.8239 & 58.88 \\ 
& \textbf{HDC} & \textbf{0.8983} & 36.50 & \textbf{0.8580} & \textbf{45.51} & \textbf{0.8331} & \textbf{57.13} \\ 
\hline

\end{tabular}
}
\caption{Quantitative results on FUGC of different semi-supervised learning methods under three data ratios. The best result is in \textbf{bold}.}
\label{FUGC}
\end{table*}

\subsection{Progressive Parameter Updates}
The teacher model is updated using an exponential moving average (EMA) of the student with main decoder:
\begin{equation}
    \theta_t \leftarrow \alpha \theta_t + (1 - \alpha) \theta_s^1,
\end{equation}
where $\alpha$ is the EMA smoothing factor, and $\theta_t$ and $\theta_s^1$ denote the weights of the teacher and the student, respectively. 

\noindent The student parameters are updated based on the total loss:
\begin{equation}
    \theta_s \leftarrow \theta_s + \eta \frac{\partial (\mathcal{L}_{total})}{\partial \theta_s}
\end{equation}
It is worth noting that the main decoder is excluded from updates by the mutual information loss to preserve its alignment with the teacher.



\section{Experiments and Results}

\subsection{Dataset}
The Fetal Ultrasound Grand Challenge (FUGC) dataset: Semi-Supervised Cervical Segmentation \cite{fugc2025} from ISBI 2025 is employed, containing 500 transvaginal ultrasound images of the uterine cervix. This dataset includes 50 labeled images and 450 unlabeled ones. For training, 40 labeled images are used, while 10 are reserved for validation. The test set consists of 90 images sourced from the challenge’s Validation Phase \cite{fugc2025}.

To further validate the model's generalization capacity, experiments are conducted on the MICCAI 2023 Grand Challenge - PSFH dataset \cite{lu2022jnu}, specifically focusing on the segmentation of the pubic symphysis (PS) and fetal head (FH). The dataset consists of 5,101 photos divided into three subsets: training (70\%), validation (10\%), and testing (20\%). Each image is linked with a precise segmentation mask that delineates the FH and PS, allowing for the successful construction and evaluation of relevant segmentation models.
\subsection{Implementation Details}

The experimental setup is based on PyTorch, utilizing an NVIDIA RTX 3090Ti with 24GB VRAM. All experiments were conducted on an Ubuntu 20.04 operating system with Python 3.8, PyTorch 1.10, and CUDA 11.3.

For the FUGC dataset, the AdamW optimizer is employed with an initial learning rate of $1\times10^{-4}$, $\beta_{CG} = 0.9$, $\beta_{MI} = 0.999$, $\epsilon = 1\times10^{-8}$, and a weight decay of 0.05. A CosineAnnealingLR scheduler is applied to ensure smooth convergence throughout training.

For the PSFH Dataset,  the same setting as DSTCT  \cite{jiang2024intrapartum} was used. Using the Stochastic Gradient Descent (SGD) optimizer set up with a momentum of 0.9 and a weight decay of 0.0001, the network training protocol involved 30,000 iterations. To accommodate semi-supervised learning paradigms, we used a batch size 16, with an equal distribution of eight tagged and eight unlabeled photos. A clustering-based learning rate technique was used to modify the training's initial learning rate of 0.01 dynamically. Data noise perturbation in the [-0.2, 0.2] range was added to improve the model's resilience to input variability
\subsection{Evaluation Metrics}
Similar to the ISBI Challenge's original assessment criteria, we validated the model's performance by employing Hausdoff Distance (HD) and Dice Score for the FUGC dataset.

Three recognized metrics—the Average Surface Distance (ASD), the 95\% Hausdorff Distance (HD95), and the Dice Similarity Coefficient (DSC)—were used to assess these models' performance on the PSFH dataset statistically. Thanks to these parameters, the accuracy and consistency of the segmentation models can be thoroughly evaluated, which is essential for confirming their clinical usefulness.

\begin{table*}[ht]
\renewcommand{\arraystretch}{1.15} 
\resizebox{\textwidth}{!}{
\begin{tabular}{|l|ccc|ccc|ccc|}
\hline
\textbf{Method} & \multicolumn{3}{c|}{\textbf{PS}} & \multicolumn{3}{c|}{\textbf{FH}} & \multicolumn{3}{c|}{\textbf{PSFH}} \\

 & \textbf{DSC↑} & \textbf{ASD↓} & $\mathbf{HD_{95}} \downarrow$ & \textbf{DSC↑} & \textbf{ASD↓} & $\mathbf{HD_{95}} \downarrow$ & \textbf{DSC↑} & \textbf{ASD↓} & $\mathbf{HD_{95}} \downarrow$ \\
\hline

SupOnly  & 0.835 & 0.793 & 3.875 & 0.897 & 1.901 & 8.868 & 0.866 & 1.347 & 6.372 \\
MT \cite{tarvainen2017mean}        & 0.803 & 0.760 & 5.143 & 0.872 & 2.912 & 13.608 & 0.837 & 1.836 & 9.376 \\
ICT \cite{verma2022interpolation}       & 0.847 & 0.813 & 3.806 & 0.896 & 1.694 & 7.191 & 0.871 & 1.254 & 5.498 \\
UAMT \cite{yu2019uncertainty}      & 0.830 & 0.801 & 4.375 & 0.889 & 1.661 & 7.296 & 0.859 & 1.231 & 5.836 \\
DAN \cite{zhang2017deep}       & 0.824 & 1.048 & 4.583 & 0.866 & 1.443 & 8.871 & 0.855 & 1.246 & 5.727 \\
DCT \cite{qiao2018deep}       & 0.801 & 1.009 & 4.960 & 0.879 & 2.081 & 9.044 & 0.840 & 1.545 & 7.002 \\
CCT \cite{Ouali_2020_CVPR}       & 0.824 & 0.596 & 4.388 & 0.886 & 2.693 & 11.604 & 0.855 & 1.645 & 7.998 \\
CPS \cite{chen2021-CPS}       & 0.829 & 0.701 & 3.788 & 0.893 & 1.385 & 5.952 & 0.864 & 1.043 & 4.870 \\
CTCT \cite{luo2022semi}      & 0.836 & 0.794 & 4.265 & 0.888 & 3.440 & 6.163 & 0.859 & 1.231 & 5.807 \\
CTCL \cite{li2022collaborative}      & 0.826 & 1.197 & 5.509 & 0.900 & 0.747 & 5.107 & 0.863 & 0.972 & 5.308 \\
S4CVnet \cite{wang2022cnn}   & 0.838 & 0.823 & 3.926 & 0.906 & 0.975 & 4.658 & 0.872 & 0.899 & 4.310 \\
DSTCT \cite{jiang2024intrapartum}   & 0.845 & \textbf{0.602} & 4.013 & 0.912 & 0.496 & 3.982 & 0.879 & 0.549 & 3.998 \\
\hline
\textbf{HDC}     & \textbf{0.850} & 0.605 & \textbf{3.911} & \textbf{0.928} & \textbf{0.466} & \textbf{3.434} & \textbf{0.889} & \textbf{0.536} & \textbf{3.673} \\
\hline

\end{tabular}
}
\centering
\caption{Quantitative results on PSFH of different semi-supervised learning methods under three data ratios. The best result is in \textbf{bold}.}
\label{PSFH}
\end{table*}

\subsection{Quantitative Results}

In this experiment, we compare HDC with existing approaches to semi-supervised learning. Unlike previous methods, HDC consists of one student and one teacher, where the student has two decoders to enhance both consistency learning and multi-task supervision.


Table \ref{FUGC} summarizes the segmentation performance of different methods under varying labeled/unlabeled data splits. HDC outperforms previous state-of-the-art methods across most settings. Specifically, under the ResNet50 backbone, our method achieves 0.8704 Dice and 38.97 HD in the 40/450 setting, slightly behind FixMatch \cite{sohn2020fixmatch}, which achieves the best HD of 38.82. Similarly, under ResNet101, FixMatch also achieves the lowest HD of 33.36, indicating its strong ability to reduce boundary errors in the high-data regime.


\begin{figure*}[h]
    \centering    \includegraphics[width=1\linewidth]{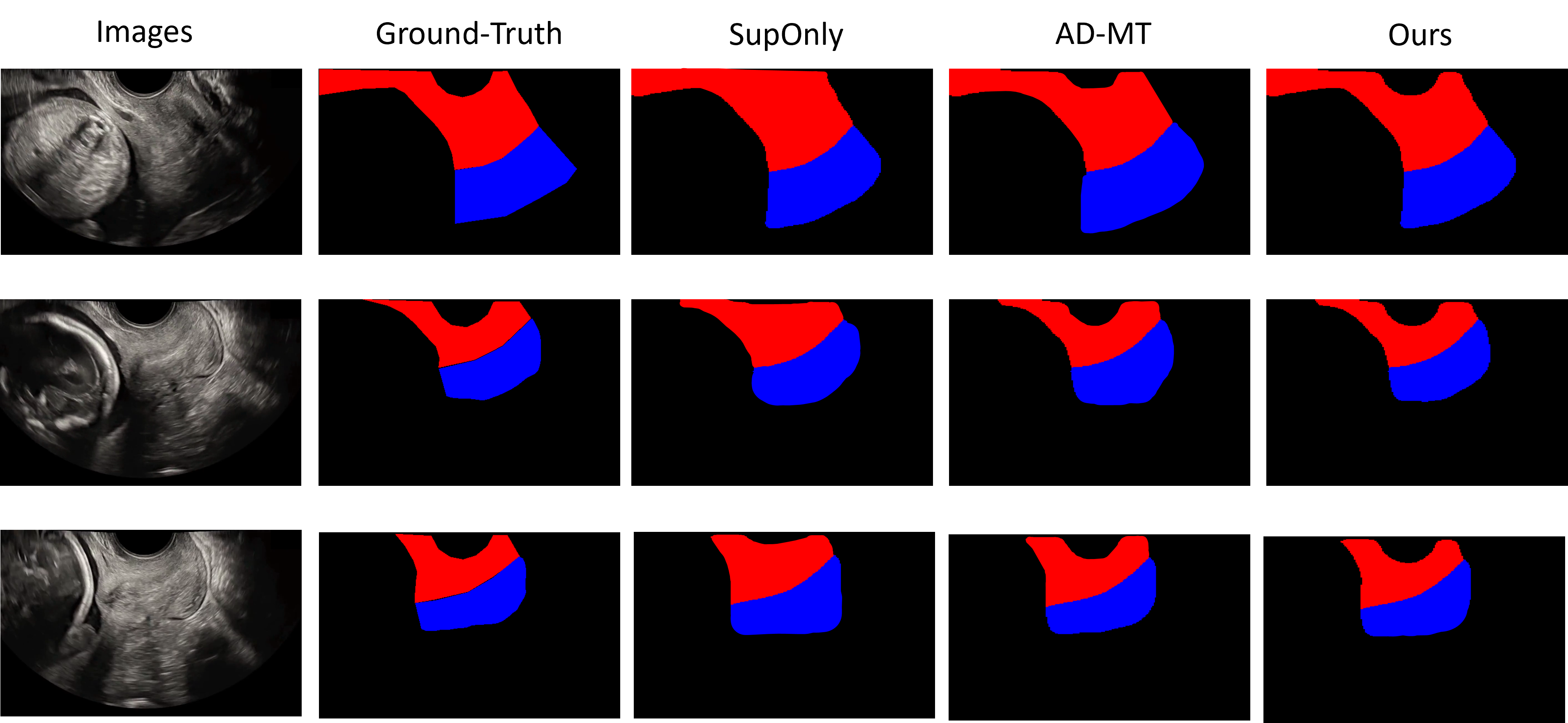}
    \caption{Qualitative comparison of different semi-supervised segmentation methods results.}
    \label{fig:full_width}
\end{figure*}

However, when the number of labeled images decreases to 20 or 10, our method achieves the best performance across both ResNet50 and ResNet101 backbones. Specifically, under the 20/470 setting, our approach achieves the highest Dice scores of 0.8503 (ResNet50) and 0.8580 (ResNet101) while also obtaining the best HD scores of 48.73 (ResNet50) and 45.51 (ResNet101). Similarly, in the most challenging 10/480 setting, our method maintains the best Dice scores of 0.8299 (ResNet50) and 0.8331 (ResNet101) and the lowest HD scores of 57.82 and 57.13, respectively.

Table \ref{PSFH} presents the quantitative outcomes on the PSFH dataset when trained with only 20\% of the total labeled training data, demonstrating the effectiveness of semi-supervised learning in leveraging unlabeled data for fetal ultrasound segmentation. Despite the significantly reduced annotation requirement, HDC consistently outperforms state-of-the-art approaches across all segmentation tasks. Specifically, model achieves a DSC of 0.850 for the PS task, surpassing DSTCT (0.845) and significantly outperforming MT (0.803). In the FH task, HDC attains a DSC of 0.928 and an HD\(_{95}\) of 3.434, improving upon DSTCT (0.920 DSC, 3.493 HD\(_{95}\)) and ICT (0.916 DSC, 7.191 HD\(_{95}\)). Similarly, for the PSFH task, our method achieves the highest DSC (0.889) and the lowest HD\(_{95}\) (3.673), outperforming all other techniques. These results highlight the strong generalization capability of our hierarchical distillation strategy, which allows the model to learn from unlabeled data more effectively.

\subsection{Qualitative Result}

The qualitative results in Fig.~\ref{fig:full_width} illustrate the segmentation performance of different methods on transvaginal ultrasound images. The supervised-only approach exhibits the weakest performance, with noticeable segmentation errors compared to the ground truth, which is expected due to its inability to leverage unlabeled data. AD-MT produces improved results, but noticeable deviations from the GT remain, particularly in boundary regions. In contrast, HDC achieves the closest segmentation to the GT, capturing finer details and reducing misclassified regions, demonstrating the effectiveness of our semi-supervised approach in leveraging unlabeled data for enhanced segmentation accuracy.

\subsection{Ablation studies}

\begin{table}[ht]
\centering

\renewcommand{\arraystretch}{1.2}
\setlength{\tabcolsep}{6pt} 

\resizebox{1\columnwidth}{!}{%
\begin{tabular}{|c|c|c|cc|}
\hline
\multicolumn{3}{|c|}{\textbf{Method}} & \multicolumn{2}{c|}{\textbf{40/450}} \\ \hline
Pix Loss & CG Loss & MI Loss & Dice  & HD   \\ 
\hline
\checkmark   &     &     & 0.8889 & 38.55 \\ 
\checkmark   & \checkmark   &     & 0.8881 & 38.72 \\ 
\checkmark  &     & \checkmark   & 0.8904 & 38.03 \\ 
\checkmark   & \checkmark   & \checkmark  & \textbf{0.8983} & \textbf{36.50} \\ 
\hline
\end{tabular}
}

\caption{Ablation study of different loss components on the 40/450 setting.}
\label{tab:ablation}

\end{table}

To analyze the contribution of different loss components, we conduct an ablation study in Table \ref{tab:ablation}. HDC integrates Pix Loss, CG Loss, and MI Loss, with the full combination yielding the highest Dice score of 0.8983 and the lowest HD of 36.50. These results indicate that each loss function contributes to better segmentation quality, and their synergy further enhances performance. HDC effectively leverages unlabeled data through consistency training, outperforming prior semi-supervised methods, particularly in low-label settings (20/470 and 10/480). The dual-decoder structure improves feature representation, leading to superior Dice scores and Hausdorff Distance reductions across most experimental settings.

\section{Conclusion}
This work presents a novel semi-supervised segmentation framework for transvaginal ultrasound imaging, addressing challenges like low contrast and shadow artifacts through adaptive consistency learning with a single-teacher architecture. By introducing Correlation Guidance Loss and Mutual Information Loss, the framework enhances feature alignment and stabilizes student learning while reducing computational complexity. Experimental results on FUGC and PSFH datasets demonstrate state-of-the-art performance with lower computational overhead than multi-teacher models, highlighting its effectiveness and generalization ability. This approach offers a promising direction for efficient medical image segmentation, with potential future applications across various imaging modalities.
\section*{Acknowledgement}

This research is supported from VNUHCM - University of Information Technology (UIT). AI VIETNAM Lab supports for providing the GPU to numerical calculate.

{
    \small
    \bibliographystyle{ieeenat_fullname}
    \bibliography{main}
}


\end{document}